# Ontology Matching Techniques: A Gold Standard Model

Alok Chauhan[1], Vijayakumar V[2], Layth Sliman[3]

[1,2]School of Computing Science & Engineering, VIT University, Chennai;
[3]EFREI, PARIS

`alok.chauhan@vit.ac.in, vijayakumar.v@vit.ac.in, layth.sliman@efrei.fr`

**Abstract.** Typically an ontology matching technique is a combination of much different type of matchers operating at various abstraction levels such as structure, semantic, syntax, instance etc. An ontology matching technique which employs matchers at all possible abstraction levels is expected to give, in general, best results in terms of precision, recall and F-measure due to improvement in matching opportunities and if we discount efficiency issues which may improve with better computing resources such as parallel processing. A gold standard ontology matching model is derived from a model classification of ontology matching techniques. A suitable metric is also defined based on gold standard ontology matching model. A review of various ontology matching techniques specified in recent research papers in the area was undertaken to categorize an ontology matching technique as per newly proposed gold standard model and a metric value for the whole group was computed. The results of the above study support proposed gold standard ontology matching model.

## 1 Introduction

Reported slowing down in speed of improvement in the field of ontology matching is the motivation behind present work [1]. It requires a fresh look into the field of ontology matching. Ontology matching can be performed at various levels, criteria, and environments leading to different kind of techniques (Fig. 1). It may be done either locally at the element level or globally at the structure level of ontologies. Matching criterion could be semantic, syntactic, terminological, structural and extensional based. Matching environment could be either context or content based. Various legal combinations of matching levels, criteria and environments give rise to whole range of concrete techniques as shown in Fig.1. As all these techniques are complimentary to each other, various matching systems use a combination of these techniques.

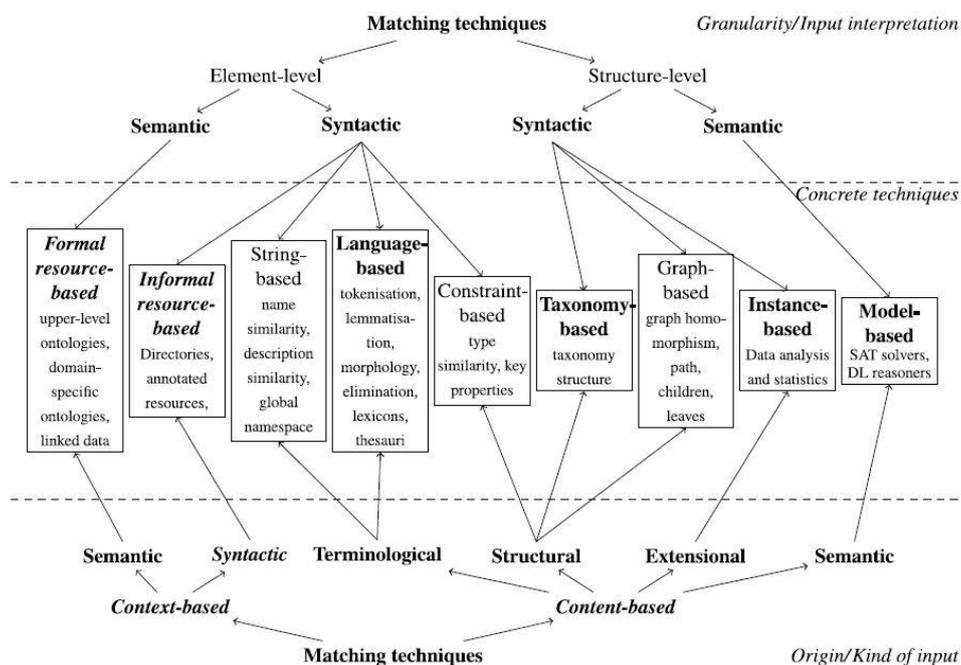

\
**Fig. 1.** A model classification of ontology matching techniques (Source: [2], p. 77)

But it leads to following questions with respect to above model classification of ontology matching techniques:

1. Are there any relationship among various matching levels, criteria and environments ?

2. Can we have any guideline to combine all the different aspects/ dimensions of ontology matching as shown in model classification, assuming it leads to some kind of synergy approximating as gold standard ?
3. Instead of arbitrary selection of concrete techniques for an ontology matching system, is there any pattern corresponding to a holistic concept of meaning ?

In the following sections, we will address these issues.

## 2    Related Work

Traditionally, almost all ontology matching surveys [3-6] focussed on classification/disintegration rather than integration, hence model proposed in this paper is a novel idea, though this integrative model would not have been possible in the absence of excellent surveys made earlier. The closest approach to the current work [7] classifies matching techniques in terms of three layers, viz., data layer, ontology layer and context layer which does have some resemblance to proposed model but goals of the two approaches are entirely different.

## 3    Gold Standard Model

Intuitively, on the basis of above mentioned classification, one might even propose a gold standard model for an ontology matching system in terms of a layered view of ontology matching criteria as shown in Fig. 2. This new model is a structured regrouping of generic ontology matching criteria in the form of a layered representation having subsumption relationship among successive layers from top to bottom. In a way the new model is derived from model classification by integrating generic matching criteria leading to a comrpehensive state-of-the-art representation of ontology matching or it looks like a compressed/folded version of model classification. As per the proposed model (Fig. 2), by "Gold Standard", we mean an ontology matching system that includes all the nine layers (Context, Content,….., Extensional).

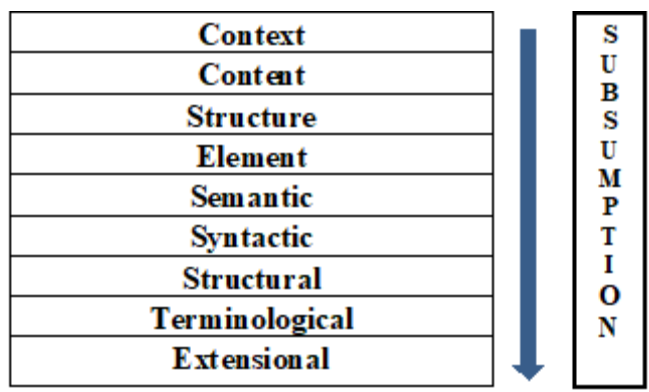

**Fig. 2.** A gold standard model for an ontology matching system

Above model represents a holistic/universal view of meaning in terms of existing constructs, though at present we do not have any methodology to support this model but an effort in this direction has been made by the authors [8] and process to propose a new foundation ontology is underway.

## 4    Coverage Metric

An appropriate metric may also be proposed to measure the coverage of ontology matching criteria by a class of ontology matching systems as per proposed model. Proposed metric called C-measure (Coverage-measure) is as follows:

$$C - Measure = \frac{\sum(frequency\ of\ ontology\ maching\ techniques)}{9 \times (number\ of\ machning\ systems)}$$

## 5    Proposed Study

The proposed comparative study of ontology matching techniques is based on the models shown in Fig. 1 and Fig. 2. For this purpose, recent research papers from www.ontologymatching.org are used as sample and titles of these papers are searched manually on the basis of appropriate keywords. The choice of keywords may seem arbitrary but the logic behind using these keywords in such a manner is to cover all the concrete techniques given in Fig. 1 and to avoid overlapping among different layers of the proposed model (Fig. 2). The whole effort is to derive an intuitive ontology matching model from the well known classification of matching techniques (Fig. 1). The approaches mentioned in these papers are classified manually on the basis of models mentioned above. Results of proposed study are found to be in accordance with proposed gold standard ontology matching model.

## 5.1 Extensional Layer

It is the innermost layer of the proposed model (Fig. 2). The online repository (http://www.ontologymatching.org/publications.html) was searched with keyword = "instance". An appropriate search word is helpful in getting more results. Overall some twenty-eight (28) research papers were reviewed in this category and chart below shows frequency distribution of various parameters/elements of the proposed gold standard model (Fig. 3, 4) [9 - 36].

| Sl No. | keyword = (instance) | context | content | structure | element | semantic | syntactic | structural | terminological | extensional |
|---|---|---|---|---|---|---|---|---|---|---|
| 1 | H. Seddiqui, R. Pratap Deb Nath, M. Aono: An efficient metric of automatic weight generation for properties in instance matching technique IJWesT, 2015 | | p | p | p | | p | p | | p |
| 2 | J. Nosner, D. Martin, P. Yeh, P. Patel-Schneider: CogMap: A Cognitive Support Approach to Property and Instance Alignment In Proceedings of ISWC, 2015 | | p | p | p | | p | p | | p |
| 3 | M. Kejriwal, Daniel Miranker: Decision-making Bias in Instance Matching Model Selection In Proceedings of ISWC, 2015 | p | p | p | | | p | p | | p |
| 4 | J. Li, Z. Wang, X. Zhang, J. Tang: Large Scale Instance Matching via Multiple Indexes and Candidate Selection Knowledge-Based Systems, 2013 | | p | p | p | | p | p | | p |
| 5 | S. Rong, X. Niu, E. Xiang, H. Wang, Q. Yang, Y. Yu: A Machine Learning Approach for Instance Matching Based on Similarity Metrics In Proceedings of ISWC, 2012. | | p | p | | | p | | | p |
| 10 | K. Zamanifar, F. Alamiyan: A New Similarity Measure for Instance Data Matching In Proceedings of CSIT, 2011. | | p | p | | | p | | | p |
| 11 | S. Wang, A. Isaac, S. Schlobach, L. van der Meij, B. Schopman: Instance-based Semantic Interoperability in the Cultural Heritage SWJ, 2011. | | p | p | | | p | | | p |
| 12 | K. Zaiss, T. Schluter, S. Conrad: Instance-Based Ontology Matching Using Different Kinds of Formalisms World Academy of Science, Engineering and Technology, 2009. | | p | p | | | p | | | p |
| 25 | T. Kirsten, A. Thor, E. Rahm: Instance-based matching of large life science ontologies In Proceedings of DILS, 2007 | | p | p | | | p | | | p |
| 26 | P. Bernstein, S. Melnik, P. Mork: Interactive Schema Translation with Instance-Level Mappings In Proceedings of VLDB (Demonstration), 2005 | | p | p | | | p | | | p |
| 27 | J. Wang, J. Wen, F. Lochovsky, W. Ma: Instance-based Schema Matching for Web Databases by Domain-specific Query Probing In Proceedings of VLDB, 2004 | | p | p | | | p | | | p |
| 28 | R. Ichise, H. Takeda, S. Honiden: Integrating Multiple Internet Directories by Instance-based Learning In Proceedings of IJCAI, 2003 | | p | p | | | p | | | p |

**Fig. 3.** Analysis of extensional layer

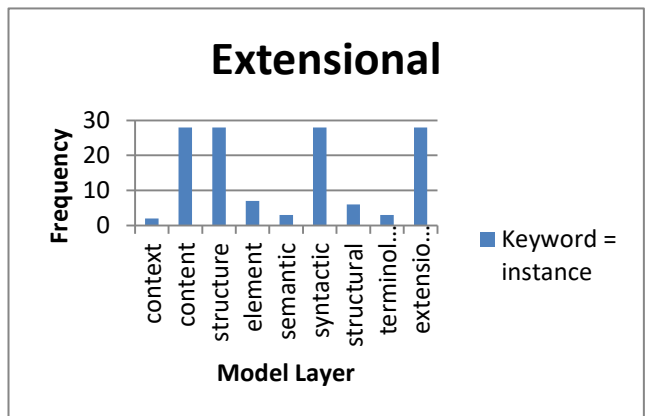

**Fig. 4.** Coverage of extensional ontology matching techniques

What we can see from above chart is that very few research papers dealing with instance-based techniques really include all the components of the proposed model, i.e., from context to extensional. One can as well see the emphasis on certain components (content, structure, syntactic) as compared to other components. The emphasis on extensional component is obvious in this case.

## 5.2 Terminological Layer

This layer is above extensional layer and online repository is searched with keywords such as "string", "lexic"ons and "thesaur"i and nineteen (19) research papers were reviewed (Fig. 5, 6) [37 - 55].

| Sl. No. | keyword = string, lexic, thesaur | context | content | structure | element | semantic | syntactic | structural | terminological | extensional |
|---|---|---|---|---|---|---|---|---|---|---|
| 1 | Y. Sun, L. Ma, S. Wang: A Comparative Evaluation of String Similarity Metrics for Ontology Alignment JOICS, 2015 | | p | | p | p | p | | p | |
| 2 | J. Wang, G. Li, D. Deng, Y. Zhang, J. Feng: Two Birds With One Stone: An Efficient Hierarchical Framework for Top-k and Threshold-based String Similarity Search In Proceedings of ICDE, 2015 | | p | | p | p | p | | p | |
| 3 | M. Cochez: Locality-Sensitive Hashing for Massive String-Based Ontology Matching In Proceedings of WI-IAT, 2014 | | p | | p | p | p | | p | |
| 4 | M. Cheatham, P. Hitzler: String Similarity Metrics for Ontology Alignment In Proceedings of ISWC, 2013 | | p | | p | p | p | | p | |
| 5 | Z. Yang, J. Yu, M. Kitsuregawa: Fast Algorithms for Top-k Approximate String Matching In Proceedings of AAAI, 2010 | | p | | p | p | p | | p | |
| 10 | S. Sen, S. Somavarapu, N.L. Sarda: Class structures and Lexical similarities of class names for ontology matching In Proceedings of ODBIS, 2007 | | p | | p | p | p | | p | |
| 11 | M. S. Chaves, V. L. Strube de Lima: Applying a Lexical Similarity Measure to Compare Portuguese Term Collections In Proceedings of SBIA, 2004 | | p | | p | p | p | | p | |
| 12 | R. Benassi, S. Bergamaschi, M. Vincini: TUCUXI: The InTelligent Hunter Agent for Concept Understanding and LeXical ChaIning In Proceedings of WI, 2004 | | p | | p | p | p | | p | |
| 16 | B. Lauser, G. Johannsen, C. Caracciolo, J. Keizer, W. van Hage, P. Mayr: Comparing human and automatic thesaurus mapping approaches in the agricultural domain In Proceedings of DC, 2008 | | p | | p | p | p | | p | |
| 17 | A. Tordai, B. Omelayenko, G. Schreiber: Thesaurus and Metadata Alignment for a Semantic E-Culture Application In Proceedings of K-CAP, 2007 | | p | | p | p | p | | p | |
| 18 | A. C. Liang, M. Sini: Mapping AGROVOC and the Chinese Agricultural Thesaurus: Definitions, tools, procedures New Review of Hypermedia and Multimedia, 2006. | | p | | p | p | p | | p | |
| 19 | A. Liang, M. Sini, C. Chun, L. Sijing, L. Wenlin, H. Chunpei, J. Keizer: The Mapping Schema from Chinese Agricultural Thesaurus to AGROVOC New review of hypermedia and multimedia, 2006 | | p | | p | p | p | | p | |

**Fig. 5.** Analysis of terminological layer

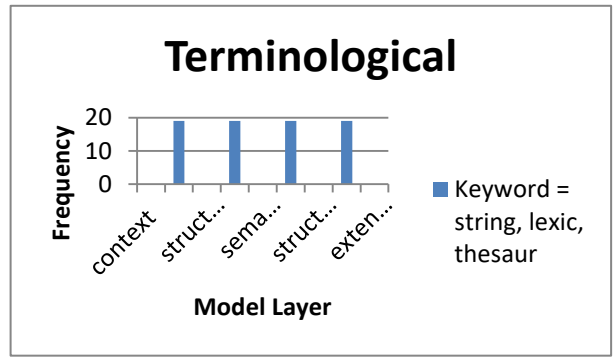

**Fig. 6.** Coverage of terminological ontology matching techniques

Though no concrete inferences can be made out of above analysis (Fig. 5), it just suggests, intuitively, the inclusion of left over components in ontology matching techniques covered under various layers from a "Gold Standard" point of view. We do find gaps in coverage of various layers in this category (Terminological) of matching techniques.

### 5.3    Structural Layer

Online repository is searched with keywords such as "structural", "constraint", "taxonom"y and "graph". Largest number of papers (42) were reviewed under this category (Fig. 7) [56 – 97]. The chart (Fig. 8) below shows the frequency distribution of various matching criteria.

Ontology matching techniques under this category seems to favor certain aspects of matching over others and this gap may be filled to achieve better results. Coverage of layers is not as good as it was in extensional techniques.

| Sl. No. | keyword = structural, costraint, taxonom, graph | context | content | structure | element | semantic | syntactic | structural | terminological | extensional |
|---|---|---|---|---|---|---|---|---|---|---|
| 1 | A. Essayeha, M. Abed: Towards ontology matching based system through terminological, structural and semantic level Procedia Computer Science, 2015 | | p | p | p | p | p | p | | |
| 2 | J.-F. Ethier, O. Dameron, V. Curcin, M. McGilchrist, R. A. Verheij, T. Arvanitis, A. Taweel, B. Delaney, A. Burgun: A unified structural/terminological interoperability framework based on LexEVS: application to TRANSFoRm JAMIA, 2013 | p | p | p | p | p | p | | | |
| 3 | V. Svatek, M. Vacura, M. Homola, J. Kluka: Mapping Structural Design Patterns in OWL to Ontological Background Models In Proceedings of K-CAP, 2013. | | p | | p | | p | p | | |
| 4 | M. Mehdi Keikha, M. Ali Nematbakhsh, B. Tork Ladani: Structural Weights In Ontology Matching IJWesT, 2013 | p | p | | | | p | p | | |
| 5 | F. Esposito, N. Fanizzi, C. d'Amato: Recovering Uncertain Mappings through Structural Validation and Aggregation with the MoTo System In Proceedings of SAC, 2010 | p | p | p | p | p | p | | | |
| 10 | G. de Melo: Not Quite the Same: Identity Constraints for theWeb of Linked Data In Proceedings of AAAI, 2013 | p | p | | | | p | p | | |
| 11 | M. Mao, Y. Peng, M. Spring: An adaptive ontology mapping approach with neural network based constraint satisfaction Journal of Web Semantics, 2010 | p | p | | | | p | p | | |
| 12 | X. Li, C. Quix, D. Kensche, S. Geisler: Automatic Schema Merging Using Mapping Constraints Among Incomplete Sources In Proceedings of CIKM, 2010. | p | p | | | | p | p | | |
| 19 | B. Berjawi, F. Duchateau, F. Favetta, M. Miquel, R. Laurini: PABench: Designing a Taxonomy and Implementing a Benchmark for Spatial Entity Matching In Proceedings of GEOProcessing, 2015 | p | p | | | | p | p | | |
| 20 | S. Raunich, E. Rahm: Target-driven merging of taxonomies with ATOM Information Systems, 2014 | p | p | | | | p | p | | |
| 29 | Y. Kitamura, S. Segawa, M. Sasajima, S. Tarumi, R. Mizoguchi: Deep Semantic Mapping between Functional Taxonomies for Interoperable Semantic Search In Proceedings of ASWC, 2008 | p | p | | | | p | p | | |
| 30 | J. J. Jung: Taxonomy alignment for interoperability between heterogeneous virtual organizations Expert Systems with Applications, 2008 | p | p | | | | p | p | | |
| 39 | A. Secer, C. Sonmez, H. Aydin: Ontology mapping using bipartite graph IJPS, 2011 | p | p | | | | p | p | | |
| 40 | P. Doshi, R. Kolli, C. Thomas: Inexact Matching of Ontology Graphs Using Expectation-Maximization Journal of Web Semantics, 2009 | p | p | | | | p | p | | p |
| 41 | J. Zhong, H. Zhu, J. Li, Y. Yu: Conceptual Graph Matching for Semantic Search In Proceedings of ICCS, 2002 | p | p | | | | p | p | | |
| 42 | S. Melnik, H. Garcia-Molina, E. Rahm: Similarity Flooding: A Versatile Graph Matching Algorithm In Proceedings of ICDE, 2002 | p | p | | | | p | p | | |

**Fig. 7.** Analysis of structural layer

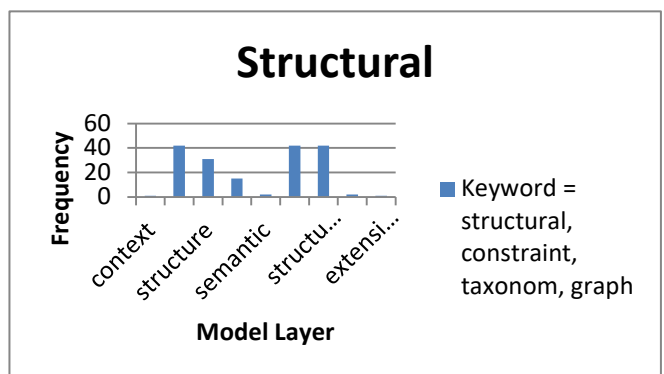

**Fig. 8.** Coverage of stuctural ontology matching techniques

### 5.4 Syntactic Layer

Keywords such as "terminological" and "annot"ated were used for the purpose of searching and it was noticed that term "terminological" is being used in a different way (such as a directory) as opposed to what we may infer from Fig. 1 (String based, Language based). Some fifteen (15) research papers were reviewed under this category (Fig. 9) [98 – 112]. The chart for this layer is shown below (Fig. 10).

| Sl. No. | Keyword = terminological, annot | context | content | structure | element | semantic | syntactic | structural | terminological | extensional |
|---|---|---|---|---|---|---|---|---|---|---|
| 1 | A. Essayeha, M. Abed: Towards ontology matching based system through terminological, structural and semantic level Procedia Computer Science, 2015 | | | p | p | p | p | p | p | |
| 2 | J.-F. Ethier, O. Dameron, V. Curcin, M. McGilchrist, R. A. Verheij, T. Arvanitis, A. Taweel, B. Delaney, A. Burgun: A unified structural/terminological interoperability framework based on LexEVS: application to TRANSFoRm JAMIA, 2013 | p | p | p | p | p | p | | | |
| 3 | B. Rance, M. Snyder, J. Lewis, O. Bodenreider: Leveraging Terminological Resources for Mapping between Rare Disease Information Sources MedInfo, 2013 | p | | | p | p | p | | p | |
| 4 | H. Saitwal, D. Qing, S. Jones, E. V. Bernstam, C. G. Chute, T. R. Johnson: Cross-terminology mapping challenges: A demonstration using medication terminological systems Journal of Biomedical Informatics, 2012. | p | | | p | p | p | | p | |
| 5 | J. Noessner, M. Niepert, C. Meilicke, H. Stuckenschmidt: Leveraging Terminological Structure for Object Reconciliation In Proceedings of ESWC, 2010 | p | | | p | p | p | | p | |
| 10 | D. Faria, E. Jimenez-Ruiz, C. Pesquita, E. Santos, F. Couto: Towards annotating potential incoherences in BioPortal mappings In Proceedings of ISWC, 2014 | p | | | p | p | p | | p | |
| 11 | M. Zhang, K. Chakrabarti: InfoGather+: Semantic Matching and Annotation Numeric and Time-Varying Attributes in Web Tables In Proceedings of SIGMOD, 2013 | p | | | p | p | p | | p | |
| 14 | K. Belhajjame, N. W. Paton, S. M. Embury, A. A. Fernandes, C. Hedeler: Feedback-Based Annotation, Selection and Refinement of Schema Mappings for Dataspaces In Proceedings of EDBT, 2010 | p | | | p | p | p | | p | |
| 15 | N. James, K. Todorov, C. Hudelot: Ontology Matching for the Semantic Annotation of Images In Proceedings of FUZZ, 2010 | p | | | p | p | p | | p | |

**Fig. 9.** Analysis of syntactic layer

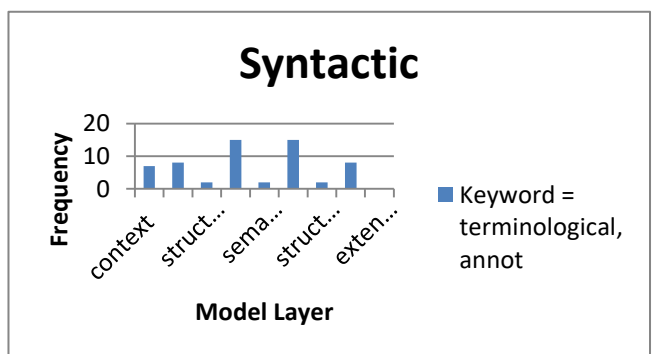

**Fig. 10.** Coverage of syntactic ontology matching techniques

### 5.5 Semantic Layer

Online repository (http://www.ontologymatching.org/publications.html) was searched with keywords "background", "upper", "context", and "sat" to get more results (22) for this layer/ category (Fig. 11) [113 – 134]. Coverage wise, we get the best results here (Fig. 12).

There is no need to search for remaining layers (Element, Structure, Content and Context) of proposed model (Fig. 2) as all the concrete techniques of standard matching technique classification model (Fig. 1) have already been covered in previous layers.

| Sl. No. | keyword = (background, upper, context, sat) | context | content | structure | element | semantic | syntactic | structural | terminological | extensional |
|---|---|---|---|---|---|---|---|---|---|---|
| 1 | A. Chauhan, V. Vijayakumarb, R. Ragalac: Towards a multi-level upper ontology-foundation ontology framework as background knowledge for ontology matching problem  Procedia Computer Science, 2015 | p | p | | p | p | p | | p | |
| 2 | V. Svatek, M. Vacura, M. Homola, J. Kluka:  Mapping Structural Design Patterns in OWL to Ontological Background Models  In Proceedings of K-CAP, 2013 | p | | | | p | p | | | |
| 3 | C. Kingkaew: Using Unstructured Documents as Background Knowledge for Ontology Matching  In Proceedings of IMLCS, 2012 | p | p | | | p | p | | p | |
| 4 | C. Quix, P. Roy, D. Kensche:  Automatic Selection of Background Knowledge for Ontology Matching  In Proceedings of SWIM, 2011 | p | p | | | p | p | p | p | |
| 5 | N. Abadie:  Schema Matching Based on Attribute Values and Background Ontology  In Proceedings of AGILE, 2009 | p | p | | | p | p | p | p | |
| 10 | A. Locoro, V. Mascardi:  A correspondence repair algorithm based onword sense disambiguation and upper ontologies  In Proceedings of KEOD, 2009 | p | p | | | p | p | | p | |
| 11 | V. Mascardi, A. Locoro, P. Rosso:  Automatic Ontology Matching Via Upper Ontologies: A Systematic Evaluation  TKDE, 2009 | p | p | p | p | p | p | | p | |
| 12 | A. Kiryakov, K. Simov, M. Dimitrov:  OntoMap: Portal for Upper-Level Ontologies  In Proceedings of FOIS, 2001 | p | p | | | p | p | p | | |
| 20 | J. Huang, J. Dang:  Context-Sensitive Ontology Matching in Electronic Business  In Electronic Business Interoperability: Concepts, Opportunities and Challenges.  IGI Global, 2010 | p | p | | | p | p | | p | |
| 21 | M. Fahada, N. Moallaa, A. Bourasa:  Towards ensuring Satisfiability of Merged Ontology  In Proceedings of ICCS, 2011 | | p | p | | p | | | | |
| 22 | M. Chan, J. Lehmann, A. Bundy:  Higher-order representation and reasoning for automated ontology evolution  In Proceedings of KEOD, 2010 | | p | p | | p | | | | |

**Fig. 11.** Analysis of semantic layer

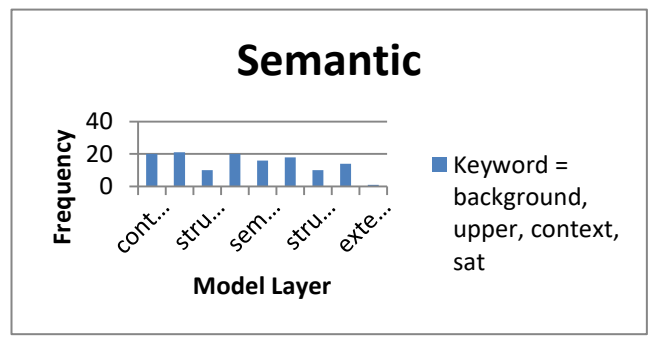

**Fig. 12.** Coverage of semantic ontology matching techniques

## 6      Results and Conclusion

Given below is metric (C-measure) computation of various layers as per above mentioned study:

| Sl. No. | Layer | C-measure |
|---|---|---|
| 1 | Extensional | **0.527778** |
| 2 | Terminological | 0.444444 |
| 3 | Structural | 0.470899 |
| 4 | Syntactic | 0.437037 |
| 5 | Semantic | **0.656566** |

**Fig. 13.** C-measure computation for various types of ontology matching techniques

It is interesting to note that concrete techniques at the beginning (Formal resource-based) and end (Instance-based and Model-based) of the spectrum (Fig. 1) fare much better with respect to proposed coverage metric, as most of the ontology matching systems falling under these categories employ most of the available ontology matching criteria. It appears as if two ends of gold standard model induce a comprehensive coverage of matching criteria due to in-built subsumption or reverse subsumption relationship. It is followed by Structural and Terminological techniques in that order (defined as per chosen keywords). As per our metric, Syntactic techniques (defined as per selected keywords) come last. Results of study reaffirm our conjecture that ontology matching techniques with higher C-measure score are best placed as per our proposed gold standard model and could be a good indicator towards the need of converging/ unifying various ontology matching efforts towards a common model as proposed by us and not just incremental improvements in individual domains. This ranking has nothing to do with individual techniques' performance as per existing vertical metrics (Precision, Recall, and F-Measure), though it has already been reported in literature that ontology matching using background knowledge does improve match result [120]. The aim of this study is to complement existing vertical metrics with newly proposed horizontal coverage metric. It is the combination of horizontal (C-measure) as well as vertical

(recall, precision and F-measure) which is expected to give much better and consistent evaluation of match results. Also, present study in a way reconfirms the standard ontology matching classification model (Fig. 1).

## 7      Limitations of Study

Due to inductive nature of hypothesis, study based on just one repository is sufficient but it can be expanded to include more resources. Similarly, manual review of papers may not be a limitation as automation may lead to compromise with quality of results.

## 8      Future Work

As future work, proposed model may be applied to popular matching systems and results may be analyzed to prove the utility of coverage metric. OAEI results may also be analyzed from this new perspective and requirement of unification of various ontology matching approaches may be assessed and emphasized.